\documentclass[conference]{IEEEtran} 
\IEEEoverridecommandlockouts

\usepackage[framemethod=TikZ]{mdframed}
\usepackage{graphicx}
\usepackage{hyperref}
\usepackage{subcaption}
\usepackage{textcomp}

\newif\iflargelesson
\largelessontrue

\iflargelesson
	\newenvironment{lesson}[1][]{
		\mdfsetup{
			frametitle={
				\tikz[baseline=(current bounding box.east), outer
				sep=0pt]
				\node[anchor=east,rectangle,fill=black!10]
				{\strut #1};}
			}
		\mdfsetup{innertopmargin=2pt,linecolor=black!15,linewidth=1.5pt,topline=true,frametitleaboveskip=\dimexpr-\ht\strutbox\relax,nobreak=true}
		\begin{mdframed}[]\relax
		}{\end{mdframed}}
\else
	\newenvironment{lesson}[1][]{
		\begin{quote}
			{\bf #1:}\relax
		}{\end{quote}}
\fi

\title{\huge Lessons Learned from Real-World Experiments with\\ DyRET: the
Dynamic Robot for Embodied Testing}

\author{
\IEEEauthorblockN{
	T{\o}nnes F. Nygaard,
	J{\o}rgen Nordmoen,
	Charles P. Martin and
	Kyrre Glette}

\IEEEauthorblockA{Department of Informatics and RITMO, University of Oslo,
Norway\\ Email: \href{mailto:tonnesfn@ifi.uio.no}{tonnesfn@ifi.uio.no}}
}

\hypersetup{pdfauthor={Nygaard et al.}, pdftitle={Lessons learned with DyRET},
pdfsubject={ICRA 2019 Learning Legged Locomotion Workshop}}

\begin{document}

\maketitle

\section{INTRODUCTION}
Robots are used in more and more complex environments, and are expected to be
able to adapt to changes and unknown situations. The easiest and quickest way to
adapt is to change the control system of the robot, but for increasingly complex
environments one should also change the body of the robot -- its morphology --
to better fit the task at hand~\cite{nygaard_WS_ICRA18}.

The theory of \textit{Embodied Cognition} states that control is not the only
source of cognition, and the body, environment, interaction between these and
the mind all contribute as cognitive
resources~\cite{wilson_frontiers13_embodied}. Taking advantage of these concepts
could lead to improved adaptivity, robustness, and
versatility~\cite{nordmoen_GECCO19}, however, executing these concepts on
real-world robots puts additional requirements on the hardware and has several
challenges when compared to learning just control~\cite{nygaard_EVOSTAR17}.

In contrast to the majority of work in \textit{Evolutionary Robotics}, Eiben
argues for real-world experiments in his ``Grand Challenges for Evolutionary
Robotics''~\cite{grandchallenges}. This requires robust hardware platforms that
are capable of repeated experiments which should at the same time be flexible
when unforeseen demands arise.

In this paper, we introduce our unique robot platform with self-adaptive
morphology. We discuss the challenges we have faced when designing it, and the
lessons learned from real-world testing and learning.

\section{THE `DyRET' ROBOT}
\begin{figure}[t] 
\vspace{2mm}
\centering
  \begin{subfigure}{0.23\textwidth}
    \includegraphics[width=\textwidth]{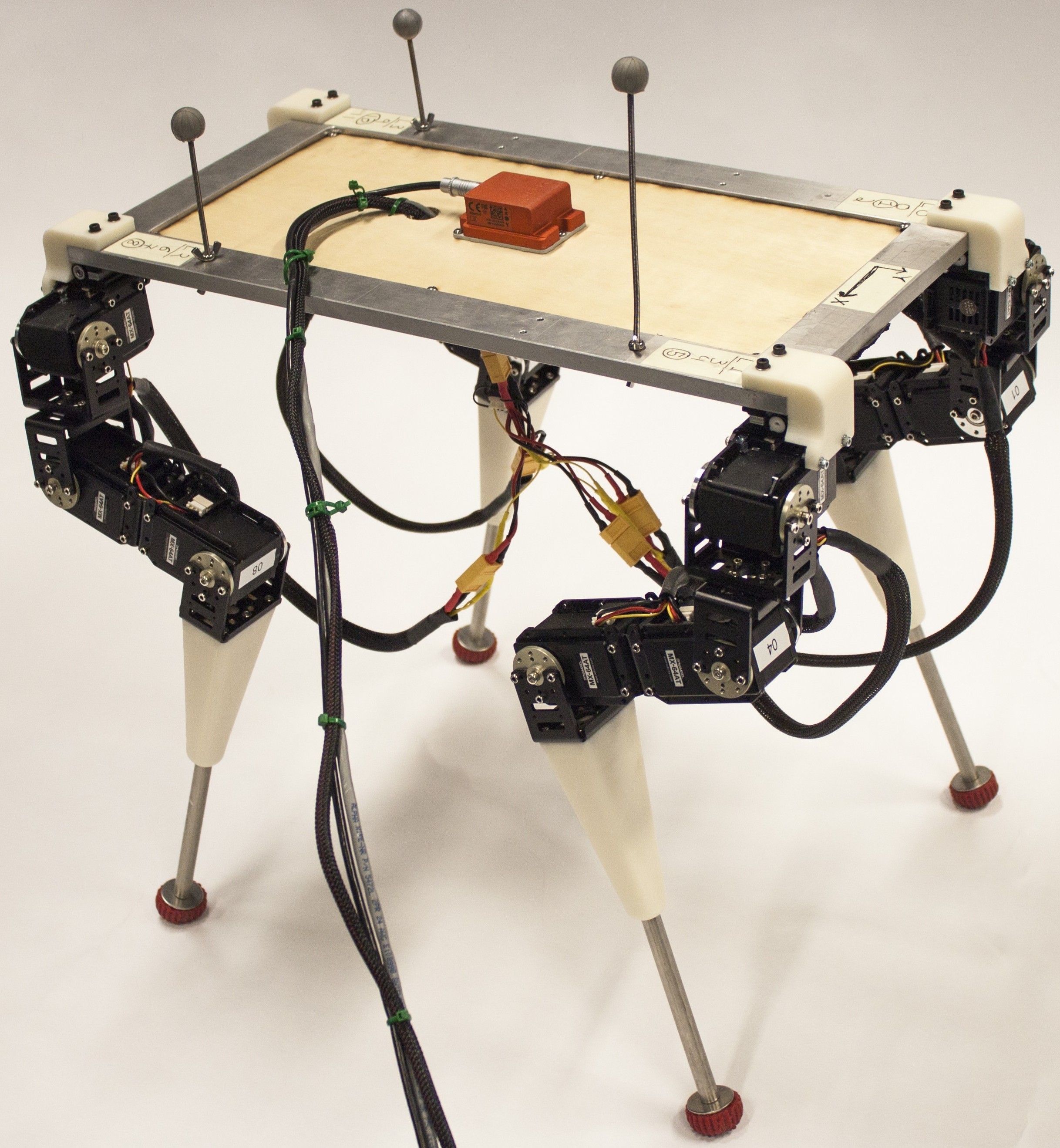}
  \end{subfigure}
  \begin{subfigure}{0.23\textwidth}
    \includegraphics[width=0.97\textwidth]{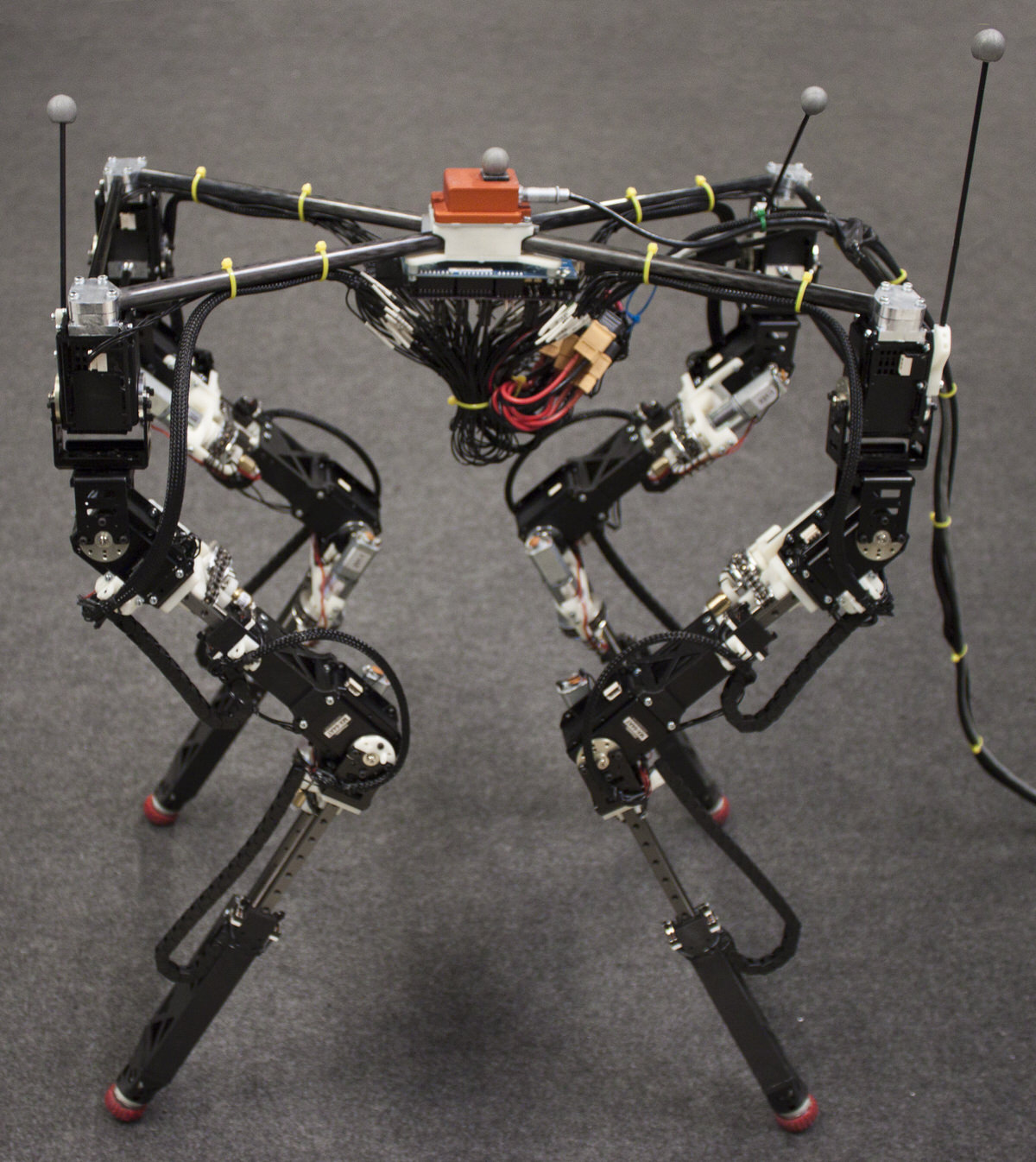}
  \end{subfigure}
  \caption{Initial version of DyRET (left) without extension legs. Latest
	version of DyRET (right) with fully extended legs.}
  \label{fig.robot}
\vspace{-2.1cm}
\end{figure}

Our robot, DyRET (Dynamic Robot for Embodied Testing), was developed to be a
platform for experiments on self-adaptive morphologies and embodied
cognition~\cite{nygaard_ICRA19}, shown in Fig.~\ref{fig.robot}. It is a fully
certified open source hardware project, and documentation, code and design files
are freely available online~\cite{github}. Since it is intended for use with
machine learning techniques it is designed to be robust, withstanding falls from
unstable gaits~\cite{nygaard_SSCI16}. It can actively reconfigure its morphology
by changing the lengths of its femur and tibia, which can be used to
mechanically gear the motors, and allow the robot to change the trade-off
between movement speed and force surplus continuously~\cite{nygaard_GECCO18}.

\section{EXPERIENCES AND CHALLENGES}
In this section, we present some key lessons we have learned when working with
DyRET. We have tried to summarize the lessons, followed by more detailed
explanations.

\begin{lesson}[Initial design considerations]
	Robustness and maintainability are more important than ease of building.
	Using \textit{rapid prototyping} and \textit{design for manufacturability} principles,
	along with exploiting COTS components is crucial in achieving an
	effective design process of a legged robot.
\end{lesson}

Legged robots are very complex systems, and anticipating all demands and
challenges early in the design process is impossible. Techniques from
\textit{rapid prototyping} allowed us to quickly get physical prototypes of the
robot, which allowed us to see and fix challenges that would be difficult to
find without having physical proof-of-concept models of the system available.
An important part of this, is to use already existing
\textit{Commercial-off-the-shelf} (COTS) components where available. This
allows us to capitalize on the work of others, and also makes it easier for
others to build or utilize lessons learned from our designs.  \textit{Desig for
manufacturability} is another important concept, and promotes adapting the
design to manufacturing considerations during the initial design process, where
they can be solved much more easily than during operation.  Making a robot that
is easy and cheap to build can be important, but our experience is that
maintainability is even more important, especially when using machine learning
that puts huge strains on the physical robot.

\begin{samepage}
\begin{lesson}[Repairs and mechanical failures]
	A good strategy for redesign is important to balance quick spot repairs
	and laborious systematic analyses of failures. Increasing the strength
	of individual parts that break is often not an effective way to do
	iterative design.
\end{lesson}
\end{samepage}

Designing parts for legged robots is always a trade-off between strength and
weight, and mechanical failures during prototyping is guaranteed. Strengthening
the part that broke can be a quick fix, but our experience is that this often
moves the problem. Both high persistent forces and sudden shock travel through
the mechanical design, and lead to failure in the next weakest link of the
chain. Reducing stress concentrations locally in a particular part can
sometimes be successful in allowing the robot to withstand a similar situation
again, however, excessive force can often lead to cascading failures throughout
the system. Having a clear strategy for when and what to do when mechanical
failures happen is important, and early on deciding on a balance between quick
spot repairs and laborious systematic analyses of failures. Once an experiment
is underway, replacing parts with similar parts might be the only option
without skewing the results, so extra efforts on failure identification during
the prototyping phase might be worth the effort. Larger cracks in the material
are often easy to identify, but deflection during operation, small fractures,
or material creep can be harder to detect.

\begin{lesson}[Controller complexity]
	Low controller complexity often entails ease of understanding with less
	dynamic results, while high complexity controllers can give good
	results, but demand more optimization. In hardware this trade-off is
	worth thinking about.
\end{lesson}

Learning legged locomotion, or a gait, is a difficult challenge. To optimize the
gait, the movement of the legs is parameterized through a gait controller. A lot
of a priori knowledge can be embedded into the controller, and result in few
parameters that are easy to optimize. Less prior knowledge requires more of the
optimization algorithm, resulting in an increased number of evaluations. The
more knowledge that is embedded, the less room there is for a varied range of
behaviors, which might be needed to adapt to new or changing tasks, environments
or the robot itself~\cite{Nordmoen18}. Finding the right complexity balance can
be very challenging, especially in real-world learning where the number of
evaluations are limited. We have successfully used a gait controller with
dynamic complexity~\cite{nygaard_EVOSTAR19}, but similar performance can be
achieved by focusing on complexity early on in the process. Another option is
using different controllers for different environments or
tasks~\cite{nordmoen2018dynamic}, for instance a complex controller when
optimizing the gait in a simulator with cheap evaluations, and a less complex
controller in the real world.

\begin{lesson}[Experiment design]
	Both the environment and the robot itself are dynamic, and changes will
	happen during operation. This can lead to biases in the experiment
	results, which have to be controlled for by proper experiment design. 
\end{lesson}

One of the key insights we have experienced after several papers worth of
real-world experiments on DyRET is how components change characteristics during
the course of experiments. Because of this gradual change it is important to
store as much information as possible so that automatic procedures can be
applied to detect differences during and after experiments. A big
difference between simulation and real-world experiments is that a
real-world experiment can never be perfectly replicated. The change in characteristics should also guide the
experiment design in the real world. Because components are expected to change
it is important to evenly test different solution so as to not bias the
experiment towards a specific solution. A concrete example is the reduction in
performance of our joints as the motors heat up. If the solutions are
always tested in the same order, this might affect the results, and give
spurious effects that cloud the results.

\begin{lesson}[Starting in the real world]
	Starting with simulation can be a quick way to get started
	learning locomotion, however, it is more difficult to transition from
	abstract simulated robots to the real world compared to going from a
	physical system to simulation.
\end{lesson}

Evaluating solutions on a physical robot system can take several seconds to
minutes, depending on gait complexity and experiment design. Evaluating on
physics simulations or with simplified models, often done in software, can give
a speedup of several orders of magnitude. This often makes simulation a
flexible and easier starting point. However, our experience with DyRET
indicates that going from a real-world robot to simulation can yield more
realistic simulation results which in turn translates to more sensible
real-world gaits after software optimization. Not basing a virtual robot on a
physical prototype makes it easier to make choices resulting in solutions that
turn out to be infeasible in the real world.  ~\cite{mouret201720}.

\section{CONCLUSION}
In this short abstract we have presented lessons learned through several years
of experimentation on the DyRET platform, which have resulted in 13 scientific
publications. From initial design considerations to the challenges, such as the
trade-off between simulated experiments and real world learning, which have
guided the choices we have taken. We hope to encourage more researchers within
the robotics community to try real-world experiments and by sharing knowledge
usually not found in publications we believe getting started with hardware could
be easier.

\bibliographystyle{IEEEtran}
\bibliography{bibliography} 

\end{document}